\pdfoutput=1

\documentclass[11pt]{article}

\usepackage[]{acl}

\usepackage{times}
\usepackage{latexsym}
\usepackage{graphicx}
\usepackage{multirow}
\usepackage{array}
\usepackage{booktabs}
\usepackage{xcolor}
\usepackage{amsmath,amssymb}
\usepackage[geometry]{ifsym}
\usepackage{color}
\usepackage{subcaption}
\usepackage{algorithmicx}
\usepackage{amsmath}
\usepackage{amssymb}
\usepackage{mathrsfs}
\usepackage{longtable}
\usepackage{amsthm}
\usepackage{amsfonts}
\usepackage{graphicx}
\usepackage{stmaryrd}
\usepackage{algorithm}
\usepackage{xr}
\usepackage{multirow}
\usepackage{times}
\usepackage{latexsym}
\usepackage{url}
\usepackage{array}
\usepackage{booktabs}
\usepackage{balance}
\usepackage{colortbl}
\usepackage{makecell}
\usepackage{soul}
\usepackage{blindtext}

\usepackage{hyperref}
\usepackage{nicefrac}       %
\usepackage{microtype}      %
\usepackage{fancyvrb}
\usepackage{cellspace, tabularx}

\usepackage{xskak}

\usepackage{wrapfig}
\usepackage{multicol}
\usepackage{xspace}
\usepackage{enumitem}
\usepackage{diagbox}

\usepackage{listings}
\usepackage{arydshln}

\usepackage{upquote}

\usepackage{threeparttable}

\usepackage[T1]{fontenc}

\usepackage[utf8]{inputenc}

\usepackage{microtype}

\usepackage{inconsolata}

\newcommand{\chessqueen}{\textsuperscript{\resizebox{0.25cm}{!}{$\BlackQueenOnWhite$}}}
\newcommand{\chessking}{\textsuperscript{\resizebox{0.25cm}{!}{$\BlackKingOnWhite$}}}
\newcommand{\chessrook}{\textsuperscript{\resizebox{0.25cm}{!}{$\BlackRookOnWhite$}}}
\newcommand{\chessknight}{\textsuperscript{\resizebox{0.25cm}{!}{$\BlackKnightOnWhite$}}}
\newcommand{\chessbishop}{\textsuperscript{\resizebox{0.25cm}{!}{$\BlackBishopOnWhite$}}}

\usepackage{amsmath,amsfonts,bm}

\def\eqref#1{equation~\ref{#1}}

\def\1{\bm{1}}

\DeclareMathAlphabet{\mathsfit}{\encodingdefault}{\sfdefault}{m}{sl}
\SetMathAlphabet{\mathsfit}{bold}{\encodingdefault}{\sfdefault}{bx}{n}

\definecolor{azure}{rgb}{0.0, 0.5, 1.0}
\definecolor{amber}{rgb}{1.0, 0.49, 0.0}
\definecolor{americanrose}{rgb}{1.0, 0.01, 0.24}
\definecolor{amethyst}{rgb}{0.6, 0.4, 0.8}
\definecolor{applegreen}{rgb}{0.55, 0.61, 0.0}
\definecolor{ballblue}{rgb}{0.13, 0.67, 0.8}
\definecolor{bazaar}{rgb}{0.6, 0.47, 0.48}
\definecolor{carminepink}{rgb}{0.88, 0.4, 0.56}

\newcommand{\gpt}{\texttt{GPT-3.5-Turbo}}
\newcommand{\gptbig}{\texttt{GPT-4}}

\title{Evaluating In-Context Learning of Libraries for Code Generation}

\stepcounter{footnote}
\author{Arkil Patel\chessking{}\thanks{\:\:Work done partly during an internship at the Allen Institute for AI.} \quad Siva Reddy\chessking{}\chessknight{}\chessrook{} \quad Dzmitry Bahdanau\chessking{}\chessknight{}\chessbishop{} \quad Pradeep Dasigi\chessqueen{}\\
	\raise3.5ex\hbox{}\chessking{}Mila and McGill University \quad \chessqueen{}Allen Institute for AI \quad
	\chessknight{}ServiceNow Research \\
        \chessrook{}Facebook CIFAR AI Chair \quad
	\chessbishop{}Canada CIFAR AI Chair
	\\
	\raise3ex\hbox{}\normalsize{\texttt{\{arkil.patel, siva.reddy, bahdanau\}@mila.quebec}} \quad \normalsize{\texttt{pradeepd@allenai.org}}  \\
}

\begin{document}
\maketitle
\begin{abstract}
Contemporary Large Language Models (LLMs) exhibit a high degree of code generation and comprehension capability. A particularly promising area is their ability to interpret code modules from unfamiliar libraries for solving user-instructed tasks. Recent work has shown that large proprietary LLMs can learn novel library usage in-context from demonstrations. These results raise several open questions: whether demonstrations of library usage is required, whether smaller (and more open) models also possess such capabilities, etc. In this work, we take a broader approach by systematically evaluating a diverse array of LLMs across three scenarios reflecting varying levels of domain specialization to understand their abilities and limitations in generating code based on libraries defined in-context. Our results show that even smaller open-source LLMs like Llama-2 and StarCoder demonstrate an adept understanding of novel code libraries based on specification presented in-context. Our findings further reveal that LLMs exhibit a surprisingly high proficiency in learning novel library modules even when provided with just natural language descriptions or raw code implementations of the functions, which are often cheaper to obtain than demonstrations. Overall, our results pave the way for harnessing LLMs in more adaptable and dynamic coding environments.
\end{abstract}

\section{Introduction}

Large Language Models (LLMs) pretrained on massive amounts of text and code data \cite{gpt4, llama2} have shown promising results across various code generation tasks \cite{codex, apps}. These models excel at both, writing programs based on natural language instructions, and at generating code to help solve downstream tasks like mathematical reasoning \cite{pal, toolformer}.


\begin{figure*}[t]
    \centering    \includegraphics[scale=0.125]{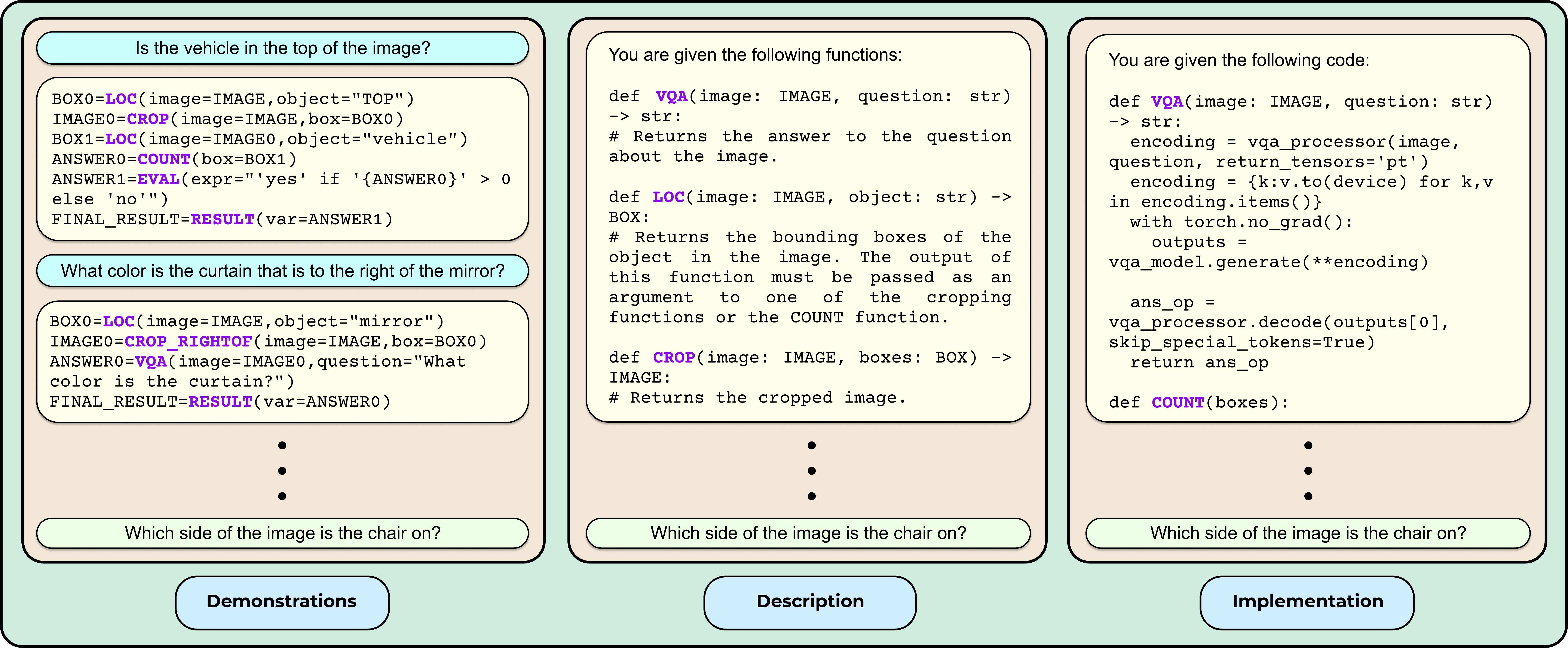}
    \caption{Illustration of the three types of in-context supervision we use to specify library functions. The examples in this figure are from the GQA dataset \cite{gqa} and the functions are from the VisProg \cite{visprog} library.} \label{fig:visprog_supervision}
\end{figure*}

As LLMs become more widely used, an important area of application is the generation of code in specialized domains. Since LLMs are expensive to finetune, in-context learning (i.e., learning from instructions and examples in the prompt) has emerged as the preferred approach for adapting LLMs to tasks and domains not seen during training. Previous works \cite{visprog, chameleon, paranjape2023art} use demonstrations provided in-context to prompt LLMs to generate code that makes calls to external task-specific library\footnote{Prior works also use the terms `API' and `tools' to refer to external programming libraries.} functions. However, there are several open questions surrounding this phenomenon: Can we use other easier-to-obtain supervision methods instead of demonstrations for the model to learn a new library? Is it possible to adapt smaller, openly accessible LLMs to generate code that uses novel libraries? Can LLMs in-context learn to use relatively uncommon programming languages? In this work, we attempt to answer these questions. We describe our evaluation framework and summarize our findings below.


\textbf{Evaluation Framework.} We design three distinct scenarios bearing real-world significance to evaluate the code generation abilities of LLMs:\\
\noindent(1) The model is \emph{constrained} to use a specific set of library functions for the task. This scenario represents code generation for software development.\\
\noindent(2) The model is required to call functions from \emph{specialized} libraries and learn their usage from the provided context. This scenario represents code generation to solve specialized tasks.\\
\noindent(3) Extending the above scenario, the model is required to generate code in a relatively uncommon programming language. This scenario evaluates whether the in-context learning ability of models is limited only to programming languages that are abundant in the pretraining data.

The underlying evaluation approach is similar across all three scenarios. We consider tasks requiring models to generate code given some natural language instruction while making calls to external library functions. Accordingly, we define \emph{novel} programming libraries or even a target programming language that the model must use to generate code. For each example in the dataset for the task, we provide a specification of the library (or language) along with the task instruction in-context to the model. We then execute the generated code and evaluate its correctness. Within this framework, we experiment with providing different types of in-context supervision for the library such as natural language descriptions or code implementations. 


\textbf{Findings.} Below, we summarize our results.\footnote{We make our code and data available at \href{https://github.com/McGill-NLP/incontext-code-generation}{https://github.com/McGill-NLP/incontext-code-generation}}\\
\noindent(1) We find that the ability to learn and use novel programming libraries is not limited to the largest proprietary models (such as GPT-4 \cite{gpt4}), but is also seen in openly-accessible models such as Llama-2 \cite{llama2} and StarCoder \cite{starcoder}.\\
\noindent(2) We show that models like GPT-4 can learn new libraries from natural language descriptions or raw code implementations of the library functions just as effectively as they can using demonstrations.\\
\noindent(3) We find that LLMs are not amenable to \emph{constrained} code generation using library functions defined in-context. They perform better when allowed to generate code without imposing constraints.\\
\noindent(4) We show that models such as GPT-4 exhibit non-trivial capability at in-context learning a new programming language from scratch.


\section{Related Work}

\setlength{\extrarowheight}{2pt}
\begin{table*}[t]
	\small{\centering
		\begin{tabular}{m{9.5em}>{\raggedright}m{7.5em}>{\centering\arraybackslash}m{4em}>{\centering\arraybackslash}m{4em}>{\centering\arraybackslash}m{5em}>{\centering\arraybackslash}m{5em}>{\centering\arraybackslash}m{5em}}
			\toprule
			\textsc{Task} & \textsc{Supervision} & \textsc{GPT-4} & \textsc{GPT-3.5} & \textsc{Llama-2} & \textsc{StarCoder} & \textsc{CodeLlama} \\
			\midrule
   
			\multirow{3}{=}{\footnotesize{GQA}} & \footnotesize{Demonstrations} &
			\footnotesize{51.10} &
			\footnotesize{51.37} &
			\footnotesize{35.75} &
			\footnotesize{32.53} &
			\footnotesize{37.35} \\
                \cline{2-7}
                & \footnotesize{Description} &
			\footnotesize{52.47} &
			\footnotesize{36.78} &
			\footnotesize{24.73} &
			\footnotesize{5.27} &
			\footnotesize{33.56} \\
                \cline{2-7}
                & \footnotesize{Implementation} &
			\footnotesize{49.11} &
			\footnotesize{44.04} &
			\footnotesize{6.85} &
			\footnotesize{27.33} &
			\footnotesize{40.85} \\
                
			\midrule

                \multirow{3}{=}{\footnotesize{NLVR}} & \footnotesize{Demonstrations} &
			\footnotesize{71.53} &
			\footnotesize{69.58} &
			\footnotesize{55.77} &
			\footnotesize{61.18} &
			\footnotesize{57.93} \\
                \cline{2-7}
                & \footnotesize{Description} &
			\footnotesize{74.85} & 
			\footnotesize{63.73} & 
			\footnotesize{54.24} & 
			\footnotesize{37.91} &
			\footnotesize{58.62} \\
                \cline{2-7}
                & \footnotesize{Implementation} &
			\footnotesize{70.49} &
			\footnotesize{57.37} &
			\footnotesize{48.91} &
			\footnotesize{51.36} &
			\footnotesize{56.44} \\

                \midrule

                \multirow{3}{=}{\footnotesize{Knowledge Tagging}} & \footnotesize{Demonstrations} &
			\footnotesize{65.93} &
			\footnotesize{62.56} &
			\footnotesize{64.67} &
			\footnotesize{58.61} &
			\footnotesize{60.92} \\
                \cline{2-7}
                & \footnotesize{Description} &
			\footnotesize{63.81} & 
			\footnotesize{42.21} & 
			\footnotesize{24.09} & 
			\footnotesize{1.54} &
			\footnotesize{29.93} \\
                \cline{2-7}
                & \footnotesize{Implementation} &
			\footnotesize{62.45} &
			\footnotesize{31.41} &
			\footnotesize{5.32} &
			\footnotesize{11.82} &
			\footnotesize{29.86} \\

                \midrule

                \multirow{3}{=}{\footnotesize{Image Editing}} & \footnotesize{Demonstrations} &
			\footnotesize{65.42} &
			\footnotesize{62.62} &
			\footnotesize{60.74} &
			\footnotesize{57.94} &
			\footnotesize{61.68} \\
                \cline{2-7}
                & \footnotesize{Description} &
			\footnotesize{62.62} & 
			\footnotesize{58.88} & 
			\footnotesize{39.25} & 
			\footnotesize{18.69} &
			\footnotesize{43.93} \\
                \cline{2-7}
                & \footnotesize{Implementation} &
			\footnotesize{64.49} &
			\footnotesize{48.59} &
			\footnotesize{32.71} &
			\footnotesize{40.19} &
			\footnotesize{45.79} \\
			
			\bottomrule
		\end{tabular}
		\caption{\label{tab:unfamiliar_lib} Performance of various LLMs at in-context learning the new VisProg library.}
	}
\end{table*}
\setlength{\extrarowheight}{0pt}

\paragraph{Evaluating Code Generation in LLMs.} There are many good benchmarks for evaluating the \emph{general} code generation abilities of language models. These include HumanEval \cite{codex}, MBPP \cite{mbpp}, APPS \cite{apps}, CodeContests \cite{alphacode}, and DS-1000 \cite{ds1000}. In this work, we focus on evaluating the capabilities of LLMs to generate code based on novel libraries defined in-context.

\paragraph{API and Tool Use.} There has recently been increased interest in studying the ability of LLMs to work with APIs and tools \cite{toolformer, gorilla, toolalpaca, toolllm, toolmanip, toolqa, apibank}. However, most of these works focus on \emph{primitive} APIs that are directly called for carrying out a specific real-world task. In this work, we focus on the problem of code generation and learning libraries (composed of complex functions) and programming languages for tasks requiring non-trivial reasoning. Unlike the above works, our work also focuses strictly on the in-context learning capability of models instead of retrieval-based or finetuning approaches. Moreover, in our work we explore questions relating to different types of in-context supervision, the capability of smaller open-source models, and the effect of enforcing constraints. Similar to our work, \citet{docprompting} and \citet{tooldoc} also explore the impact of providing documentation of APIs instead of demonstrations. While \citet{docprompting} focused on a general retrieval-based pipeline to improve performance, we study in-context learning of new libraries for specific tasks. Our findings on providing function documentations in-context are similar to \citet{tooldoc}. However, unlike their work, we show that models can achieve high performance even without including partial examples of how to call the library functions. Moreover, unexplored by these works, we also study the impact of providing raw implementations of library functions and the task of learning a new programming language from just a description of keywords.


\section{Learning Novel Libraries}\label{sec:unfamiliar_lib}

Many practical applications of code generation require learning new libraries or frameworks. Moreover, programmers need to constantly adapt to changes in existing frameworks. Motivated by these practical needs, we study the ability of LLMs to adapt to novel programming libraries in-context. \citet{visprog} showed that LLMs can be used to solve computer vision tasks by learning how to call the functions of a novel library based on demonstrations provided in-context. We study this phenomenon in more detail with a wider range of models and different types of supervision.

\subsection{Experimental Setup}

\paragraph{Tasks.} We experiment with $4$ vision-language tasks as used by \citet{visprog}. Apart from maintaining consistency to previous work, these tasks aptly demonstrate practical use cases of in-context library learning.

\noindent (1) \textbf{GQA} \cite{gqa} is a compositional visual question answering task.

\noindent (2) \textbf{NLVR} \cite{nlvr} is a reasoning task over image pairs. Given a pair of images, the task is to determine whether the corresponding statement about the images is true or false.

\noindent (3) \textbf{Knowledge Tagging} \cite{visprog} involves tagging objects in a given image.

\noindent (4) \textbf{Image Editing} \cite{visprog} involves editing a given image using computer vision tools based on a given instruction.

\paragraph{Library.} We use the custom `VisProg' library defined by \citet{visprog}, which has $20$ modules that can be called in a Python program to solve the above tasks. These modules include functionalities such as manipulating images with computer vision models, querying LLMs, etc. 

\paragraph{In-context supervision.} We experiment with three different types of supervision explained below and illustrated in Figure \ref{fig:visprog_supervision}.\\
\noindent\underline{\emph{Demonstrations}}: We provide examples of instructions paired with corresponding programs that illustrate how to call the VisProg library functions. We used the same set of exemplars as was used by \citet{visprog}. We provided a total of 20 in-context demonstrations in the prompt.\\
\noindent\underline{\emph{Description}}: We provided natural language descriptions of the library functions. We include the documentation of every function in VisProg in a Python docstring format. For each function, we specify its name, return type, names and types of its arguments, and describe its functionality.\\
\noindent\underline{\emph{Implementation}}: We directly provide the Python implementations of all the functions in VisProg.

We experiment with providing \emph{descriptions} and \emph{implementations} of library functions because both are arguably less expensive to obtain compared to paired demonstrations. Descriptions can generally be easily obtained from the documentation of the library. Moreover, since the functions of a library are already implemented in the underlying programming language, the implementation data is also readily available.

It is important to note that for the \emph{description} and \emph{implementation} types of supervision, the model has no exposure to any programs or \emph{any} information about the kinds of instructions in the domain. We provide information for \emph{all} the functions in the library, and the model needs to determine which functions it needs to use to solve the given example.

\paragraph{Models.} We experiment with GPT-4 \cite{gpt4}, GPT-3.5-Turbo \cite{gpt3, instructgpt}, LLaMA-2-70B \cite{llama2}, StarCoderPlus \cite{starcoder}, and CodeLlama \cite{codellama}.

\begin{figure}[t]
	\centering
	\includegraphics[scale=0.11, trim=0 0 0 0, clip]{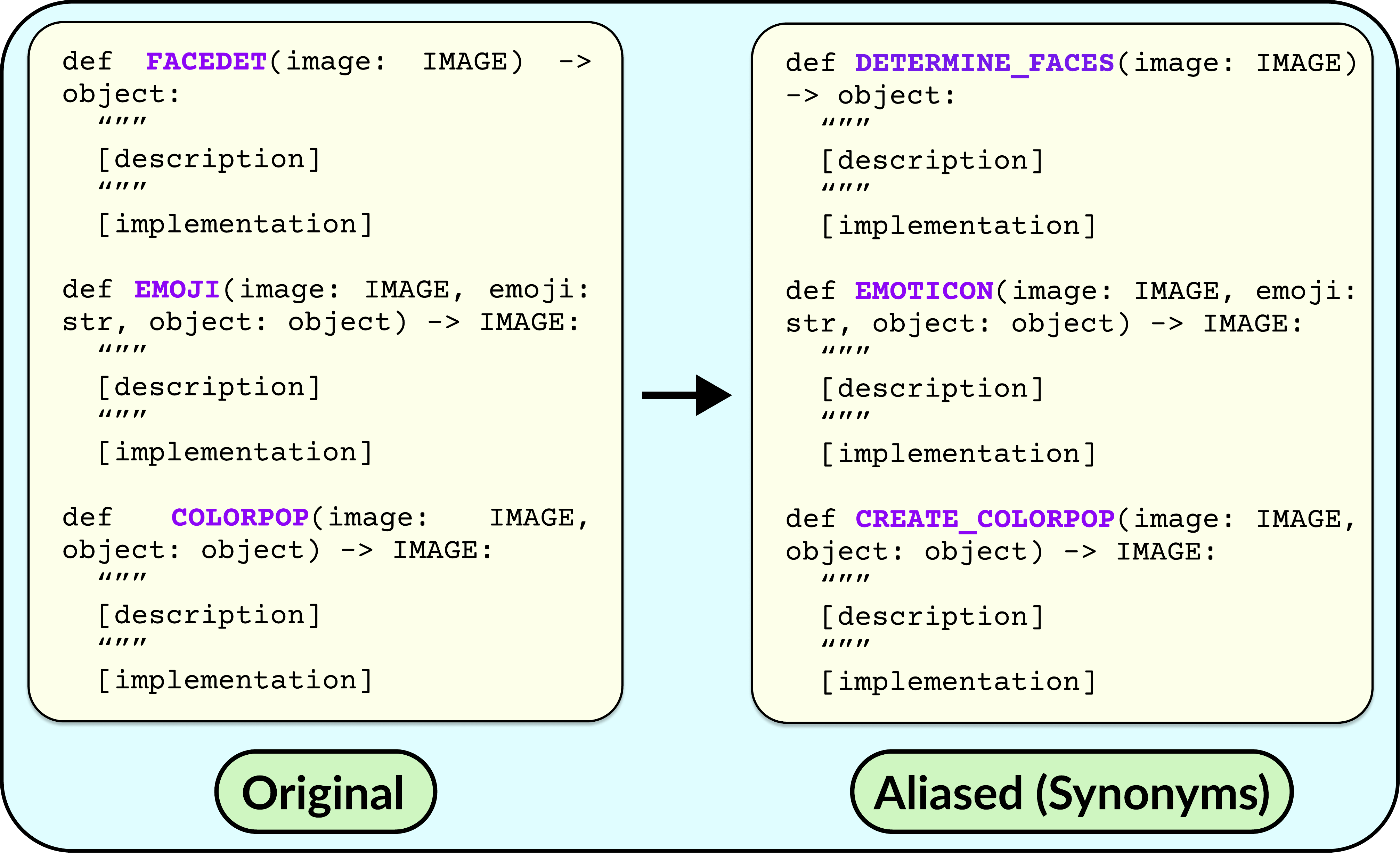}
	\caption{\label{fig:visprog_syn_ex} Illustration of aliasing the function names in VisProg with synonymous words.}
\end{figure}

\paragraph{Metrics.} We measure correctness of usage of the provided novel library by evaluating the downstream performance for the task.\footnote{We designed a code interpreter to execute the generated programs which would fail if and only if the model does not correctly use the VisProg library functions.} For GQA and NLVR, we measure accuracy of the answer against the dataset. For Knowledge Tagging, we follow \citet{visprog} and measure tagging performance via precision (fraction of predicted boxes that are correct) and recall (fraction of ground truth objects that are correctly predicted). These metrics require both the predicted bounding box and the corresponding tag to be correct. We use an Intersection-over-Union (IoU) threshold of 0.5. We summarize the performance by calculating the F1 score, which is simply the harmonic mean of
the average precision and recall across instructions. For Image Editing, we calculate correctness by carrying out manual evaluation to check if the executed program yields the correct image based on the instruction in the example.


\subsection{Results}

The results for all models across all datasets can be seen in Table \ref{tab:unfamiliar_lib}.\footnote{We also illustrate the common types of errors made by \gptbig{} in Appendix \ref{app:errors}.}


\begin{table}[t]
	\small{\centering
		\begin{tabular}{p{3.5em}p{2em}>{\centering\arraybackslash}m{1em}>{\centering\arraybackslash}m{1.5em}>{\centering\arraybackslash}m{1.5em}>{\centering\arraybackslash}m{1em}>{\centering\arraybackslash}m{1.5em}>{\centering\arraybackslash}m{1.5em}}
			\toprule
			& & \multicolumn{3}{c}{\textbf{GQA}} &\multicolumn{3}{c}{\textbf{KnowTag}} \\ 
			\cmidrule(lr){3-5}\cmidrule(lr){6-8}
			 & & OG & SYN & RAN & OG & SYN & RAN\\
			\midrule
                \multirow{3}{=}{\footnotesize{GPT-4}} & Demo & 51.1  & 49.2  & 48.6 & 65.9  & 64.9 & 60.2 \\
                & Desc & 52.5 & 48.8  & 47.8 & 63.8 & 60.3 & 58.9 \\
                & Imple & 49.1  & 48.5  & 48.3 & 62.5 & 59.5 & 57.6 \\
                \midrule
                \multirow{3}{=}{\footnotesize{GPT-3.5}} & Demo & 51.4  & 33.7  & 32.4 & 62.6 & 59.8 & 54.1 \\
                & Desc & 36.8  & 33.8  & 31.7  & 42.2 & 26.3 & 24.0 \\
                & Imple & 44.0  & 32.9  & 27.1 & 31.4 & 18.6 & 15.3 \\
			\midrule
                \multirow{3}{=}{\footnotesize{Llama-2}} & Demo & 35.8  & 33.9  & 28.7  & 64.7  & 65.8 & 55.3 \\
                & Desc & 24.7  & 25.8  & 18.6 & 24.1  & 21.3 & 16.5 \\
                & Imple & 6.9  & 5.8  & 1.5  & 5.3  & 6.7 & 1.9 \\
			\bottomrule
		\end{tabular}
		\caption{\label{tab:visprog_alias}Performance ($\uparrow$) of LLMs at in-context learning VisProg when function names are aliased. OG: original, SYN: synonymous word, RAN: random string.}
	}
\end{table}

\paragraph{Models learn to use new libraries.} We see that most models are able to learn the novel library from demonstrations across all tasks. This shows that this ability to adapt to novel code modules in-context is not limited to the biggest proprietery LLMs but is also exhibited to a good extent by openly accessible smaller models.

\paragraph{Models can learn from description and code.} We find that most models exhibit non-trivial ability\footnote{The performance is significantly more than the zero-shot baselines reported in Appendix \ref{app:zero_shot}.} of learning from just \emph{descriptions} and \emph{implementation} of the functions in the novel library without any exposure to demonstrations. Remarkably, GPT-4's performance with descriptions and implementation is comparable to that from demonstrations. This clearly shows that the best contemporary LLMs are able to understand novel code modules and use them without needing any demonstrations. However, for models apart from GPT-4, providing demonstrations still remains the best form of supervision across all tasks. In Appendix \ref{app:visprog_unpaired}, we show that providing random programs (i.e., not paired with input) along with the descriptions or implementations improves performance.

\paragraph{Effect of pretraining.} We observe that the data on which the models have been pretrained on influences the choice of supervision that best suits them. For instance, LLaMA models, which have been primarily trained on text with comparatively lesser code pretraining \cite{llama2, codellama}, show a much higher ability to learn from descriptions compared to code implementations. This is opposite for the StarCoder, which has been primarily pretrained on code. We see that StarCoder is better able to leverage implementation supervision, despite being a smaller model. However, its performance is extremely low when provided with Natural Language descriptions.

\begin{figure}[t]
	\centering
	\includegraphics[scale=0.45, trim=8 10 10 5, clip]{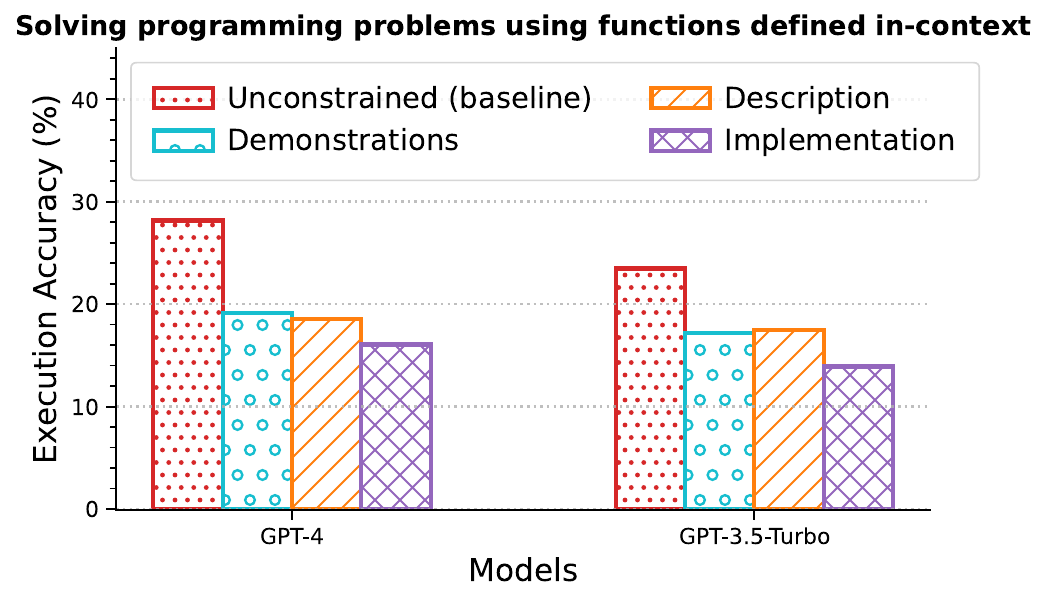}
	\caption{\label{fig:nl2p_main} Performance ($\uparrow$) of models at solving NL2Python programming problems in our curated dataset using functions defined in-context.}
\end{figure}

\paragraph{Impact of Aliasing Function Names in VisProg.} Since most LLMs that we experiment with do not disclose their pretraining data, it is unclear whether they achieve high performance because they are already familiar with the VisProg library. Hence, we also experiment with aliasing the function names in the library with synonymous words (see illustration in Figure \ref{fig:visprog_syn_ex}) or random strings. The results are provided in Table \ref{tab:visprog_alias}. There is a very clear drop in performance with aliasing (even with synonymous words) for GPT-3.5, indicating that some level of familiarity with the VisProg language biases model performance. For Llama-2, the performance with synonymous function names is similar to the original performance while there is a significant drop with random strings, indicating that Llama models rely on the semantics of function names. GPT-4 is quite robust to both types of aliasing.

\section{Constrained Generation Using Functions Defined In-Context}\label{sec:familiar_lib}
Developers often have constraints on the functions they can use. For example, coding for software projects may require using functions within the current repository. Motivated by this use case, we examine a scenario where the language model is constrained (using natural language instructions) to use specific library functions presented in-context. Note that in this scenario, it is possible for the model to generate the semantically correct code without using the functions specified in-context. Hence, we additionally evaluate whether the generated code uses the context-defined functions.


\begin{table}[t]
	\small{\centering
		\begin{tabular}{p{6em}>{\centering\arraybackslash}m{2.5em}>{\centering\arraybackslash}m{2.75em}>{\centering\arraybackslash}m{2.5em}>{\centering\arraybackslash}m{2.75em}}
			\toprule
			&\multicolumn{2}{c}{\textbf{GPT-4}} &\multicolumn{2}{c}{\textbf{GPT-3.5}} \\ 
			\cmidrule(lr){2-3}\cmidrule(lr){4-5}
			 & All & Correct & All & Correct\\
			\midrule
			Demonstrations & 89.92 & 90.11 & 20.25 & 17.23 \\
			Description & 83.22 & 82.74 & 27.03 & 18.39  \\
                Implementation & 85.38 & 84.76 & 19.13 & 15.79  \\
			\bottomrule
		\end{tabular}
		\caption{\label{tab:func_usage}Quantifying percentage usage of function(s) defined in-context in the code predicted by the model for solving the programming problem. In the \emph{All} column, we calculate the percentage of model predictions that used the function defined in-context while under the \emph{Correct} column, we calculate the percentage of correct predictions that used the context-defined function.}
	}
\end{table}

\subsection{Experimental Setup}

\paragraph{Task.} We consider the Natural Language to Python task, i.e., generating Python code from natural language instructions. We curated a dataset consisting of a total of 15000 examples sampled from the CodeContests \cite{alphacode} and APPS \cite{apps} datasets. Apart from being popular benchmarks for the NL2Code task, CodeContests and APPS consist of problems whose solutions are not just a few lines of code, but require implementation of complex logic using existing library functions or need the creation of specific user-defined functions.

\paragraph{Library.} We create a custom library of functions. The procedure used to create this library and gather the associated data is described in Section \ref{sec:lib_creation}.

\paragraph{In-context supervision.} For this scenario, we again experiment with three different types of supervision formats. For \emph{demonstrations}, we provide $5$ exemplars of a simple instruction and corresponding program (that uses the function) pairs in-context. For \emph{description}, we provide natural language documentation of the functions. For each function, we specify its name, return type, the name and types of its arguments and a brief description of its functionality. For \emph{implementation}, we directly provide the Python implementations of the functions. Example prompts with the \emph{demonstrations} and \emph{description} types of supervision are provided in Figure \ref{fig:prompt_constrained_demos} and Figure \ref{fig:prompt_constrained_desc} in the Appendix.

\paragraph{Models.} We experiment with GPT-4 and GPT-3.5-Turbo. Preliminary experiments showed that solving such difficult programming problems in-context is still very challenging for smaller openly-accessible models, so we only provide results for the latest GPT models.

\begin{figure}[t]
	\centering
	\includegraphics[scale=0.13, trim=0 0 0 0, clip]{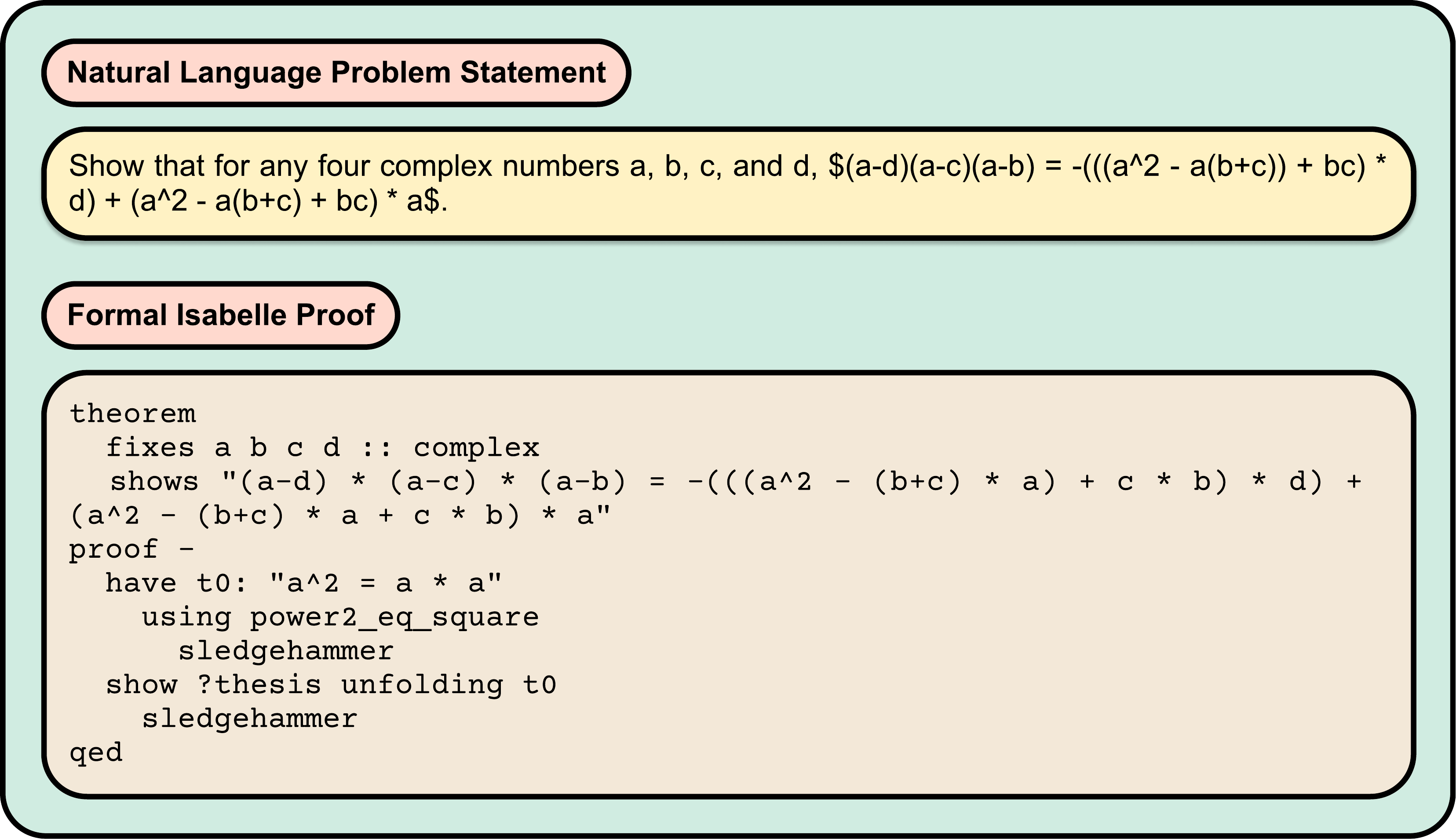}
	\caption{\label{fig:isabelle_ex} An example of the automated theorem proving task using the Isabelle language.}
\end{figure}

\paragraph{Metrics.} We evaluate using two metrics. (1) \emph{Execution Accuracy} measures the correctness of the generated program by executing it over a set of test cases. (2) \emph{Function Usage \%} measures how often does the model actually use the functions in its solution. To evaluate whether a predicted code $C$ uses a particular function $f$, we check the semantic consistency\footnote{Note that the mere presence of a function in a program does not guarantee that the output of the program depends on that function. Accordingly, exactly verifying this dependency is a hard problem. In this work, we use our semantic consistency check as an approximation.} of the predicted code before and after we replace the calls to $f$ with calls to another function $f'$ with same data types for the arguments as $f$ and a fixed return value. If the replacement alters the semantics of $C$ (i.e., the outputs for a range of different inputs are different before and after replacing $f$ with $f'$), we conclude that the code $C$ uses the function $f$.

\subsection{Creating a Library of Functions}\label{sec:lib_creation}

We create our data and library by using examples from the CodeContests \cite{alphacode} and APPS \cite{apps} datasets. Every example in these datasets consists of an instruction, candidate code solutions (in Python), and test cases to automatically evaluate the generated code. All the candidate solutions use standard Python code and libraries which the LLMs are most probably already familiar with. So, we create a novel library by extracting two types of functions that are used in the candidate solutions and aliasing the names:

\paragraph{(1) Existing Library Functions.} These are the functions that are defined in existing Python libraries such as NumPy, Pandas, etc. For a particular example selected from the above two datasets, we extract all such functions that are used in the candidate solutions for that example and alias their names with some random string.

\paragraph{(2) User-defined Functions.} These are the functions that are custom defined in the candidate solutions (i.e., with the \emph{def} keyword). Again, we extract all such functions from the candidate solutions and alias their names with random strings.

Our generated dataset consists of 12000 examples in which the model will be constrained to use existing standard library functions (the function names will be aliased) and an additional 3000 examples in which the model will be constrained to use user-defined functions extracted from the candidate solutions in the above-mentioned datasets.

\subsubsection{Obtaining Specification Data}

For the functions that we extracted from the solutions of the above-mentioned datasets, we need to obtain specification data (such as demonstrations and descriptions for the functions) which we will provide to the model in-context.

\paragraph{Obtaining Descriptions.} This is straightforward for the case of \emph{existing python library functions}: we simply scrape the API documentation of the library. For the \emph{user-defined functions}, we prompt GPT-4 with the function definition and ask it to generate the API documentation for the function. We ensure correctness of this description by a cyclic evaluation process detailed in Appendix \ref{app:cyclic}.

\paragraph{Obtaining Demonstrations.} We prompt GPT-4 with the function and its description and ask it to generate five creative examples of instruction-program pairs where the instruction is a natural language query and the program (2-3 lines) solves the instruction using the function.

\begin{figure}[t]
	\centering
	\includegraphics[scale=0.11, trim=0 0 0 0, clip]{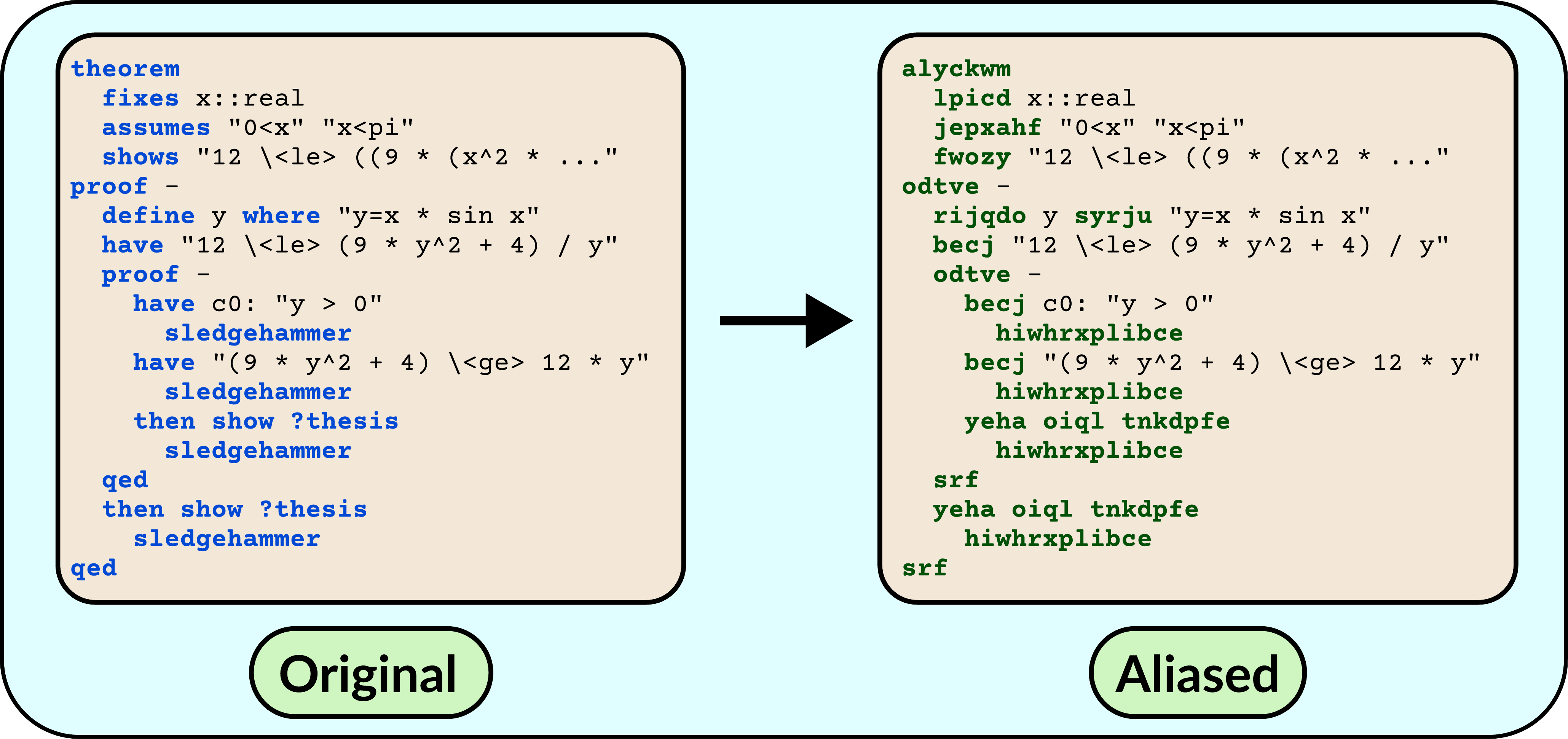}
	\caption{\label{fig:isabelle_alias_ex} Illustration of aliasing the Isabelle language.}
\end{figure}

\paragraph{Obtaining Implementations.} This is straightforward for the case of `user-defined functions' as we directly have the Python implementations of the functions. For the library functions, we extract the implementation from the source code if available. For certain libraries, the source is implemented in a language other than Python (for e.g., in C). For such cases, we prompt GPT-4 with the function and its description and ask it to generate its Python implementation. We again check the correctness of this implementation using a similar cyclic evaluation process as the one used for descriptions.

\subsection{Results}

\paragraph{Models perform worse when constrained.} The main results are provided in Figure \ref{fig:nl2p_main}. For both models, GPT-4 and GPT-3.5-Turbo, the execution accuracy (independent of function usage) decreases significantly when we constrain them to use specific functions in their prediction. This is a bit surprising and clearly shows that these models do not respond very well to constraints being put on them for tasks where they can generate the correct code without explicit supervision of any library functions. We hypothesize that the models have memorized how to program with standard library functions and find it hard to learn to use new ones with similar functionality.

\paragraph{Robustness to format of supervision.} As seen in Figure \ref{fig:nl2p_main}, the performance of both models is quite similar across the three different types of supervision we experiment with: \emph{demonstrations}, \emph{descriptions}, and \emph{implementation}. This shows that models are able to understand constraints equally well irrespective of the format of supervision used for the functions provided in-context.

\paragraph{GPT-4 follows constraints better.} In Table \ref{tab:func_usage}, we report the \emph{Function Usage \%} for both models across different prompt settings. We observe that while GPT-4 actually follows the function usage constraints defined in-context, GPT-3.5 mostly ignores them. The magnitude of the difference in Function Usage \% between the two models is quite high, indicating the difference in the quality of instruction-following behaviour between them.

\section{Learning a New Programming Language}\label{sec:unfamiliar_lang}

\begin{figure}[t]
	\centering
	\includegraphics[scale=0.4, trim=8 10 10 5, clip]{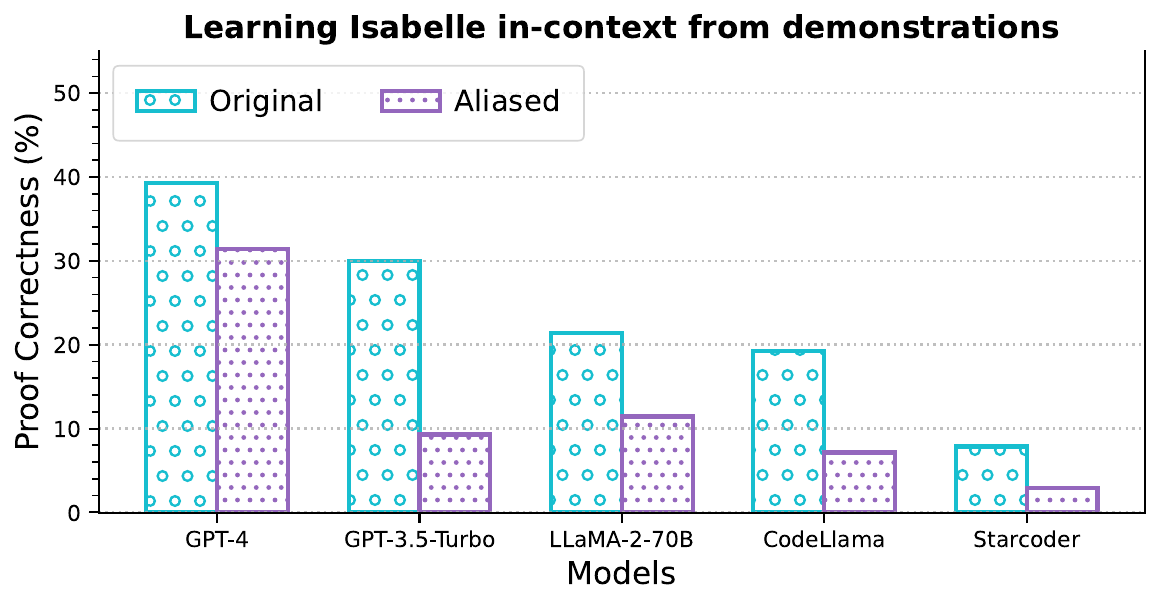}
	\caption{\label{fig:isabelle_demos} Performance ($\uparrow$) of various LLMs at learning Isabelle in-context from demonstrations (with and without aliasing with random strings).}
\end{figure}

In this section, we study the ability of LLMs to learn a new and unfamiliar programming language in-context. There are many diverse applications and tasks that use niche domain-specific-languages (DSLs) \cite{grammar_prompting,dsp, semantic_machines}. Ideally, we would want to adapt a general LLM to solve these tasks without the added overhead of re-training or finetuning the model. Hence, it is important to contextualize how good current LLMs are at learning unfamiliar languages using just in-context learning.

\subsection{Experimental Setup}

\paragraph{Task.} We consider the task of automated theorem provin,g which has immense practical relevance. We generate proofs with the Isabelle language \cite{isabelle} for examples in the miniF2F dataset \cite{minif2f}. Figure \ref{fig:isabelle_ex} provides an example of the task. We chose to work with Isabelle because it is a programming language that does not have much paired data existing on the internet. It is also relatively small (in terms of number of keywords) which makes it possible to describe it in-context. We focus on the algebra subset of MATH problems \cite{math_dataset} in the miniF2F dataset, comprising a total of 140 examples. We wish to evaluate how well the model is able to in-context learn the unfamiliar Isabelle language. The prompt provided to the model consists of some supervision about the Isabelle language (described in detail later). This is followed by a formal statement written in Isabelle and an informal proof sketch written in Natural Language \cite{dsp}. Given this prompt, the goal of the model is to generate the formal proof for the statement in the unfamiliar Isabelle language.

\paragraph{In-context supervision.} We experiment with two types of supervision formats: \emph{demonstrations} and \emph{description}. For demonstrations, we provide $8$ exemplars in the prompt, where each example consists of a formal statement in Isabelle, the corresponding informal proof in Natural Language, and the formal proof in Isabelle. Figure \ref{fig:prompt_isabelle_demos} in the Appendix provides an example of this prompt type. For description, we provide a Natural Language description of every keyword in the Isabelle language in the prompt. Note that since we are focusing on a limited set of algebra problems, the language can be represented by just $12$ keywords. Figure \ref{fig:prompt_isabelle_desc_alias} in the Appendix provides an example of this prompt type (with aliasing, which is described below).

\paragraph{Aliasing.} Since the data on which contemporary LLMs have been trained on is not widely known, it is unclear to what extent they might already be familiar with Isabelle. Hence we experiment with aliasing all the keywords of the language with a different random string. Figure \ref{fig:isabelle_alias_ex} provides an illustration of aliasing an isabelle program.

\paragraph{Models.} We experiment with GPT-4, GPT-3.5-Turbo, LLaMA-2 70B, CodeLlama, and StarCoder.

\paragraph{Metrics.} We measure \emph{proof correctness}, calculated as the fraction of examples for which the model generated the correct formal proof. Following \citet{dsp}, for a given example, we evaluate correctness of the generated proof using the Isabelle proof checker and the Sledgehammer proof automation tool \cite{sledgehammer}.

\begin{table}[t]
	\small{\centering
		\begin{tabular}{p{6.5em}>{\centering\arraybackslash}m{2.5em}>{\centering\arraybackslash}m{2.5em}>{\centering\arraybackslash}m{2.5em}>{\centering\arraybackslash}m{2.5em}}
			\toprule
			&\multicolumn{2}{c}{\textbf{GPT-4}} &\multicolumn{2}{c}{\textbf{GPT-3.5-Turbo}} \\ 
			\cmidrule(lr){2-3}\cmidrule(lr){4-5}
			 & Original & Aliased & Original & Aliased\\
			\midrule
			Demonstrations & 39.3 & 31.4 & 30.0 & 9.29 \\
			Description & 15.0 & 7.86 & 7.86 & 0.00  \\
			\bottomrule
		\end{tabular}
		\caption{\label{tab:isabelle_desc}Performance ($\uparrow$) of GPT models at learning Isabelle in-context from only a description of keywords.}
	}
\end{table}

\subsection{Results}

\paragraph{Learning Isabelle from demonstrations.} The performance of all models at learning Isabelle in-context from demonstrations can be seen in Figure \ref{fig:isabelle_demos}. We see that all models show a good amount of capability at learning the language in-context using demonstrations. This is very interesting because the task requires models to combine non-trivial reasoning ability with the syntax of this new language. The performance decreases when we alias the keywords of the library indicating that familiarity with the language may be partly responsible for the performance. Nevertheless, our aliasing results clearly show that models can learn completely new programming languages in-context to some extent.

\paragraph{Learning Isabelle from descriptions.} The results for learning Isabelle from just a description of keywords are provided in Table \ref{tab:isabelle_desc}. Only the GPT models exhibited non-trivial performance in this setting. Models show a fair amount of performance at learning Isabelle in this setting without any exposure to examples. However, considering the high zero-shot performance of GPT-4 (see Appendix \ref{app:zero_shot}), it is very likely that models have been exposed to the language during training. The performance diminishes under aliasing but is still quite significant for GPT-4, indicating some preliminary ability of learning a new language from just its description. These results are particularly relevant because the space of demonstrations for a new programming language would grow exponentially.

\begin{figure}[t]
	\centering
	\includegraphics[scale=0.38, trim=2 2 0 2, clip]{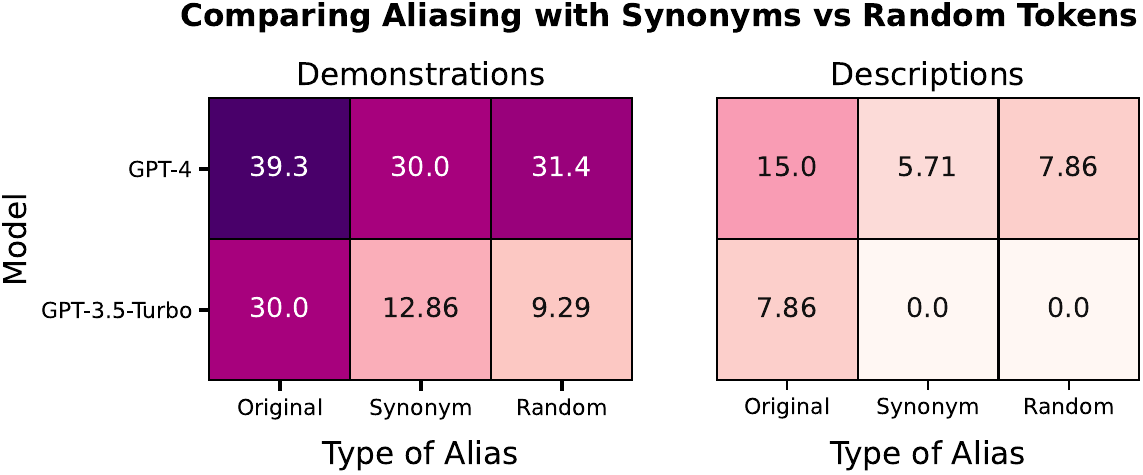}
	\caption{\label{fig:isabelle_syn} Performance ($\uparrow$) of models at learning Isabelle in-context when the keywords are aliased with a synonymous word instead of a random string.}
\end{figure}

\paragraph{Effect of aliasing with synonyms.} We experiment with aliasing the keywords of Isabelle with synonymous words instead of random strings. The results are provided in Figure \ref{fig:isabelle_syn}. We observe that the model performs similarly with synonymous aliasing as it does with random aliasing. This is quite surprising and shows that models do not depend much on the semantics of the keywords while learning a new programming language in-context.

\begin{figure}[t]
	\centering
	\includegraphics[scale=0.4, trim=8 5 10 5, clip]{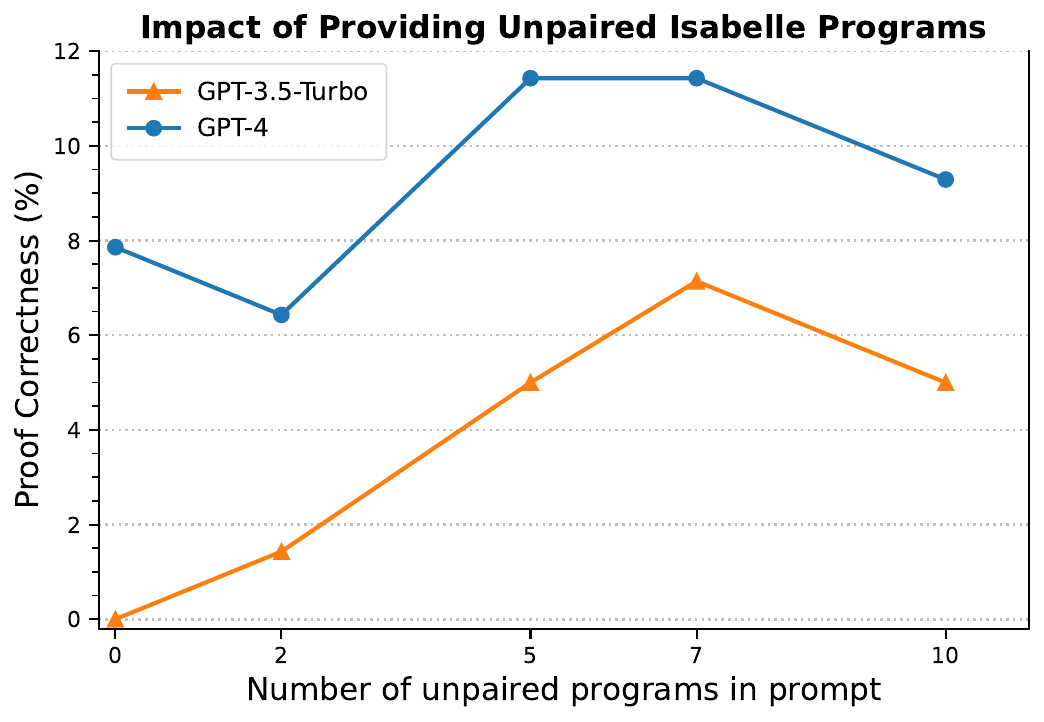}
	\caption{\label{fig:isabelle_unpaired_examples} Performance ($\uparrow$) of models at learning Isabelle in-context from just descriptions (aliased) increases with exposure to unpaired Isabelle programs.}
\end{figure}

\paragraph{Providing unpaired programs.} We also experimented with providing unpaired programs (i.e., examples of Isabelle proofs without corresponding inputs) along with the description in the aliasing experiments. Since paired supervision is expensive to obtain, our goal was to check whether models can effectively leverage random instances of programs in the unfamiliar language to better learn it. The results for the GPT models can be seen in Figure \ref{fig:isabelle_unpaired_examples}. We observe a trend that the ability to learn the unfamiliar language increases with exposure to examples of programs in the language (even without the corresponding inputs). However, the performance saturates after providing 5-7 unpaired programs, possibly due to overfitting.

\section{Discussion}

In this work, we investigated the abilities of LLMs to in-context learn novel programming libraries and languages. Below we discuss the main takeaways.

We observed that the strongest LLMs can learn novel libraries from just their natural language descriptions as well as the underlying programming language implementations. This holds promise for adapting LLMs rapidly for different applications without requiring any effort for obtaining paired demonstrations data. Moreover, we found that that the ability to acquire novel libraries in-context is not limited to proprietary LLMs. Smaller, openly-accessible models such as Llama-2 and StarCoder also exhibit a significantly high ability at learning code modules without requiring any finetuning.

We noticed that the choice of supervision provided in-context (between natural language descriptions and programming language implementations) is crucial in determining performance, especially for smaller, openly-accessible models. This finding is beneficial for resource-constrained scenarios, where we can only use particular LLMs and need to determine the type of supervision to provide, or we only have a particular type of supervision available and need to select the best model suited for it.

We found that LLMs degrade in performance when constrained to use particular libraries in scenarios where they can generate code by themselves. In such cases, LLMs have a very strong bias to generate code based on their priors and find it difficult to use the library functions provided in-context.

Finally, we observed that LLMs show a preliminary but promising ability to learn new programming languages from scratch based on in-context demonstrations or description of the language.
\section*{Limitations}

We experiment with three different scenarios of learning a novel programming library or language. However, for each scenario, we show results only for a single domain (or library/language). While we believe that our experiments are sufficient to draw the conclusions presented in the paper, in the future, we will consider more domains to further strengthen our results.

The \emph{description} and \emph{implementation} prompts of supervision for section \ref{sec:unfamiliar_lib} and \ref{sec:unfamiliar_lang} were manually crafted by one of the authors and verified by the others. This inherently inserts some bias into our results. However, note that our study was exploratory in nature and our main focus was on \emph{analyzing} the in-context learning abilities of LLMs in learning novel libraries rather than propose general supervision approaches for these tasks.

There are limitations associated with our automatic data creation procedure for obtaining description and implementation data for library functions. The created dataset may be slightly biased because the descriptions and implementations are created by an LLM itself. However, since code generation for competitive programming problems is an objective reasoning task, we believe that the effect of this bias in our dataset will be minimal.

Since this work involves automatic data creation (without manual check for each data point) using LLMs, it is possible (although unlikely) that the model generates unsafe responses. Also, a significant portion of the data is built on existing code generation and reasoning datasets, so, the biases in these datasets will transfer to our evaluation suite.

This work includes results for OpenAI models, which may not be directly reproducible.

\section*{Acknowledgments}

We thank our colleagues at the Allen Institute for AI and at Mila and McGill University for helpful discussions and for providing valuable feedback. Arkil is partly supported by the Canada Graduate Scholarship – Master's (CGS-M) funded by the Natural Sciences and Engineering Research Council of Canada (NSERC).

\bibliography{custom}

\clearpage
\newpage
\appendix

\section{Implementation Details}\label{app:imple}

Experiments using \gpt{} and \gptbig{} (version $0613$) were performed using the OpenAI API\footnote{\href{https://platform.openai.com/}{https://platform.openai.com/}}. All other experiments were done on $8$ NVIDIA A100 GPUs with $80$ GB memory. Our code is implemented in PyTorch \cite{pytorch} and makes use of the HuggingFace Transformers library \cite{huggingface}. For experiments with certain open-source models such as Llama-2, we use Huggingface Text-Generation-Inference.\footnote{\href{https://github.com/huggingface/text-generation-inference}{https://github.com/huggingface/text-generation-inference}}

For experiments with VisProg in Section \ref{sec:unfamiliar_lib}, the context lengths of the prompt are approximately $2000$ tokens for \emph{demonstrations}, $500$ tokens for \emph{description}, and $2400$ tokens for \emph{implementation}.

We use Portal-to-ISAbelle\footnote{\href{https://github.com/albertqjiang/Portal-to-ISAbelle}{https://github.com/albertqjiang/Portal-to-ISAbelle}} \cite{pisa} to evaluate Isabelle proofs. We work with Isabelle2021 and use the default Sledgehammer configuration, including a 120-second timeout.

\begin{figure}[h]
    \centering    \includegraphics[scale=0.095]{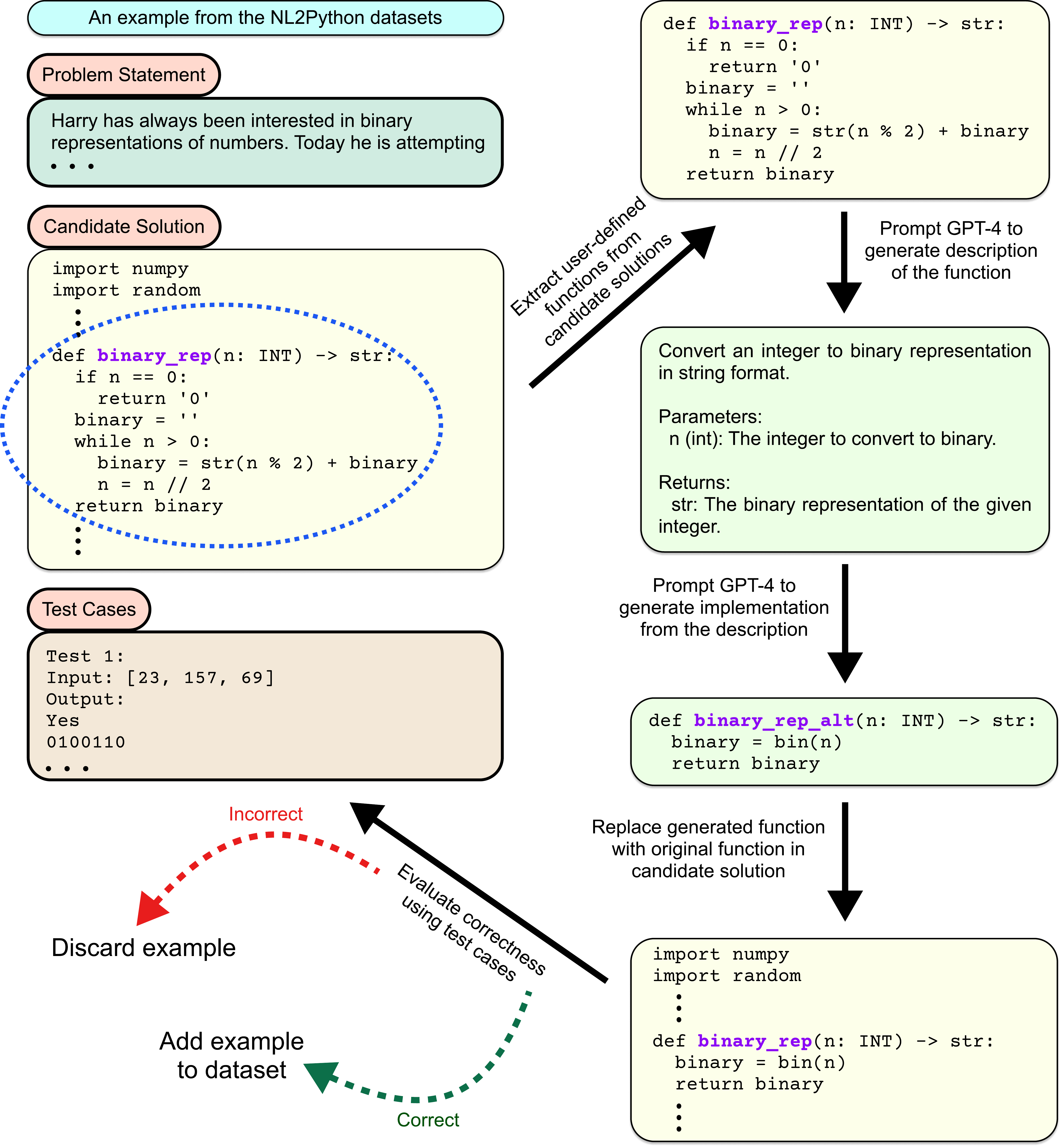}
    \caption{An illustration of the cyclic evaluation process used to obtain correct descriptions of \emph{user-defined} functions at scale.} \label{fig:cyclic_illustration}
\end{figure}

Below, we provide additional details for the datasets used in Section \ref{sec:unfamiliar_lib}.

\noindent (1) \textbf{GQA} \cite{gqa} is a compositional, multi-step visual question answering task. For each example, the task is to answer the question associated with a given image. We experiment on the test set which has 1460 examples.

\noindent (2) \textbf{NLVR} \cite{nlvr} is a reasoning task over image pairs. In each example, a pair of images is provided and the task is to determine whether the corresponding statement about the images is true or false. We experiment on the test set which has 6967 examples.

\noindent (3) \textbf{Knowledge Tagging} (KnowTag) \cite{visprog} involves identifying people and objects in a given image. The dataset has a total of 100 examples.

\noindent (4) \textbf{Image Editing} \cite{visprog} involves editing a given image using computer vision tools based on a given instruction. The dataset has a total 107 examples.\\
\section{Cyclic Evaluation Process to Obtain Correct Descriptions and Implementations}\label{app:cyclic}

\begin{table}[t]
	\small{\centering
		\begin{tabular}{>{\centering\arraybackslash}m{6em}>{\centering\arraybackslash}m{9em}}
			\toprule
			\textbf{Model} & \textbf{Zero-shot Accuracy} \\ 
			\midrule
			GPT-4 & 28.58 \\
			GPT-3.5-turbo & 11.17 \\
                Llama-2 & 1.78 \\
                CodeLlama & 1.23 \\
                StarCoder & 0.62 \\
			\bottomrule
		\end{tabular}
		\caption{\label{tab:gqa_zero_shot}Zero-shot accuracies of all models for the GQA dataset.}
	}
\end{table}

\begin{table}[t]
	\small{\centering
		\begin{tabular}{>{\centering\arraybackslash}m{6em}>{\centering\arraybackslash}m{12em}}
			\toprule
			\textbf{Model} & \textbf{Zero-shot Proof Correctness} \\ 
			\midrule
			GPT-4 & 27.86 \\
			GPT-3.5-turbo & 2.14 \\
                Llama-2 & 10.71 \\
                CodeLlama & 8.57 \\
                StarCoder & 0.71 \\
			\bottomrule
		\end{tabular}
		\caption{\label{tab:isabelle_zero_shot}Zero-shot proof correctness of all models for the algebra subset of MATH problems in the miniF2F dataset.}
	}
\end{table}

Here, we describe the cyclic evaluation process we used to automatically ensure the correctness of the function descriptions generated from GPT-4. The process is illustrated in Figure \ref{fig:cyclic_illustration}. We prompt GPT-4 with just the generated documentation of the function and ask it to generate the python implementation of the function. We then evaluate the semantic equivalence of this generated function with that of the extracted function. We can check semantic equivalence by replacing the extracted function definition with this generated function definition in the corresponding candidate solution and evaluating correctness using the test cases. 

We use a similar process for obtaining correct implementations as well. We append the generated implementation of the function in the candidate solution and evaluate correctness using the test cases.
\section{Additional Results}

\subsection{Zero-shot Baselines} \label{app:zero_shot}

To better contextualize our results presented in the main paper, we evaluate models zero-shot on the tasks without providing any library or language specification.

\paragraph{Learning new library.} We attempted to prompt the models to generate code without specifying any library or providing any examples. However, we were unable to automatically execute any of the model generations because of errors such as improper file path (which the model would assume by itself) or we ran into library dependency issues. Hence we tried to check if the models could solve examples zero-shot without generating code.

Since GQA is a question-answering dataset, we measure the zero-shot accuracies by prompting the model with only the question and asking it to guess the answer. The zero-shot accuracies for all models are provided in Table \ref{tab:gqa_zero_shot}. NLVR is a binary classification dataset. We achieved the maximum zero-shot performance of 51.1\% by prompting the models to always answer `True' (i.e., majority class baseline). Since knowledge tagging and image editing datasets require complex image editing, and object localization and classification, their examples cannot be solved without generating code and their zero-shot baseline can be considered to be 0\%.

\paragraph{Learning new language.} We prompt the models to zero-shot generate the Isabelle proof for each example in the algebra subset of MATH problems in the miniF2F dataset. The zero-shot proof correctness for all models is provided in Table \ref{tab:isabelle_zero_shot}.

\begin{figure}[t]
	\centering
	\includegraphics[scale=0.38, trim=8 10 10 5, clip]{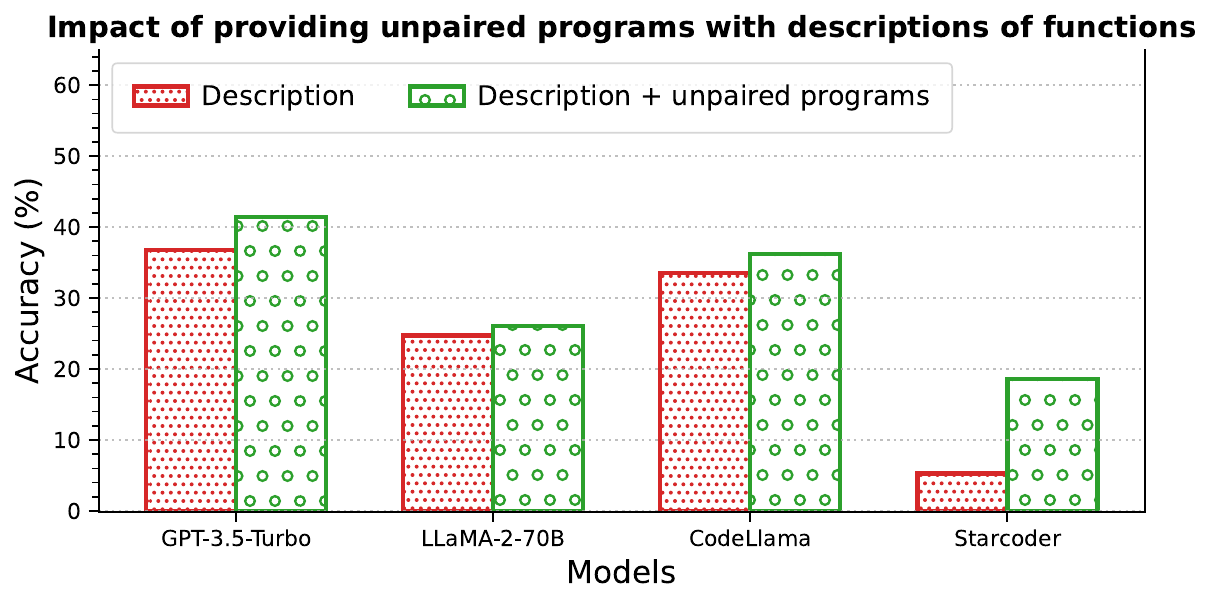}
	\caption{\label{fig:visprog_unpaired_desc} Performance ($\uparrow$) of various LLMs on GQA dataset when $10$ unpaired VisProg programs are provided along with the descriptions of functions.}
\end{figure}

\begin{figure}[t]
	\centering
	\includegraphics[scale=0.38, trim=8 10 10 5, clip]{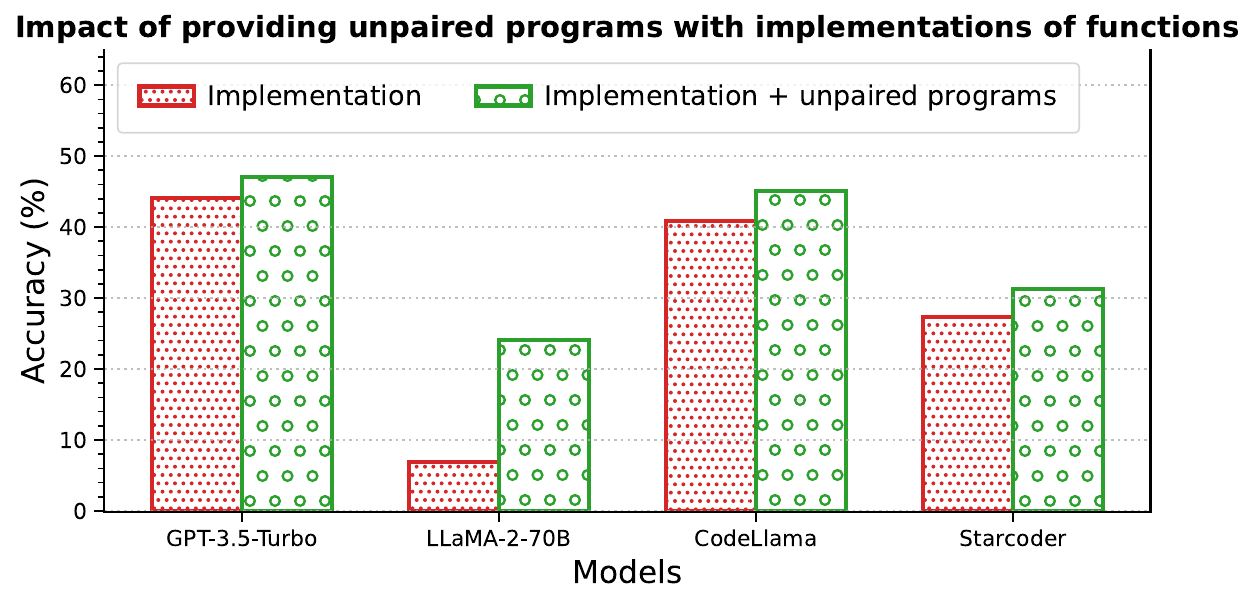}
	\caption{\label{fig:visprog_unpaired_imple} Performance ($\uparrow$) of various LLMs on GQA dataset when $10$ unpaired VisProg programs are provided along with the implementations of functions.}
\end{figure}


\subsection{Impact of Providing Unpaired Programs for VisProg} \label{app:visprog_unpaired}

In Section \ref{sec:unfamiliar_lib}, we observed that no model except GPT-4 could match the demonstrations performance with just descriptions or implementation. We wanted to check whether models can effectively leverage unpaired programs (i.e., just random programs without corresponding inputs) to perform better when provided with description or implementation specification. We experiment with GPT-3.5-turbo, Llama-2, CodeLlama, and StarCoder over the GQA dataset. We use the same prompt as we used for \emph{description} and \emph{implementation} in Section \ref{sec:unfamiliar_lib}, appended with $10$ random VisProg programs for GQA. The results for \emph{description} and \emph{implementation} are provided in Figure \ref{fig:visprog_unpaired_desc} and Figure \ref{fig:visprog_unpaired_imple} respectively. We observe improvement in all models for both types of supervision. The improvements are particularly noticeable when the model performance without the unpaired programs was extremely low, as is the case for StarCoder with \emph{descriptions} and Llama-2 with \emph{implementations}.
\section{Examples of Errors}
\label{app:errors}

In Table \ref{tab:errors_visprog} and Table \ref{tab:errors_isabelle}, we provide examples of common error patterns that we observed for predictions made by \gptbig{} while learning VisProg and Isabelle respectively. We were unable to identify concrete error patterns for our experiments in Section \ref{sec:familiar_lib}; most prediction failures were due to incorrect application of the function provided in-context.

\setlength{\extrarowheight}{0pt}
\begin{table*}[t]
	\small{\centering
		\begin{tabular}{m{45em}}
			\toprule
			  \textsc{Examples of Errors in Generating VisProg}\\
			\midrule
			\scriptsize{\textbf{\textcolor{brown}{Problem:}} Does the mat have the same color as the sky?
            \newline \textbf{\textcolor{violet}{Generated Program:}} 
            \newline
            \texttt{temp = VQA(image = IMAGE, question = \textcolor{red}{``Does the mat have the same color as the sky?"})}
            \newline
            \texttt{FINAL\_RESULT = RESULT(var = temp)}}\\
            \midrule
            \scriptsize{\textbf{\textcolor{brown}{Problem:}} Hide the faces of Nicole Kidman and Brad Pitt with ;) and 8)
            \newline \textbf{\textcolor{violet}{Generated Program:}} 
            \newline
            \texttt{faces = FACEDET(image = IMAGE)}
            \newline
            \texttt{nicole\_kidman = SELECT(image = IMAGE, object = faces, query = `Nicole Kidman', category = None)}
            \newline
            \texttt{temp\_img = EMOJI(image = \textcolor{red}{nicole\_kidman}, emoji = `winking\_face')}
            \newline
            \texttt{brad\_pitt = SELECT(image = IMAGE, object = faces, query = `Brad Pitt', category = None)}
            \newline
            \texttt{temp\_img = EMOJI(image = \textcolor{red}{brad\_pitt}, emoji = `smiling\_face\_with\_sunglasses')}
            \newline
            \texttt{FINAL\_RESULT = RESULT(var = temp\_img)}}\\
	    \bottomrule
		\end{tabular}
		\caption{\label{tab:errors_visprog} Illustrations of errors made by \gptbig{} while learning VisProg. In the top example, the model fails to correctly utilize the helper functions in the library such as `\texttt{LOC}' and `\texttt{CROP}' to break down the complex question and focus on specific image locations, thereby putting the onus completely on the \texttt{VQA} engine, which fails. The bottom example illustrates a failure in understanding the data types of the arguments as well as the values returned by the functions in the library (the \textcolor{red}{highlighted} variables are of type `\texttt{OBJECT}' and should have been passed to the `\texttt{object}' argument instead of `\texttt{image}').}
	}
\end{table*}
\setlength{\extrarowheight}{0pt}

\setlength{\extrarowheight}{0pt}
\begin{table*}[t]
	\small{\centering
		\begin{tabular}{m{45em}}
			\toprule
			  \textsc{Examples of Errors in Generating Isabelle}\\
			\midrule
			\scriptsize{\textbf{\textcolor{brown}{Problem:}} Show that for any complex number a, $(a-10)(a+11) = a^2 + a - 110$.
            \newline \textbf{\textcolor{violet}{Generated Proof:}} 
            \newline
            \texttt{have $(a-10) * (a+11) = a^2 - 10*a + 11*a - 10 *11$ sledgehammer}
            \newline
            \texttt{\textcolor{red}{hence} $(a-10) * (a+11) = a^2 + a - 10 * 11$ sledgehammer}
            \newline
            \texttt{\textcolor{red}{hence} $(a-10) * (a+11) = a^2 + a - 110$ sledgehammer}
            \newline
            \texttt{finally show ?thesis sledgehammer}
            \newline
            \texttt{qed}}\\
            \midrule
            \scriptsize{\textbf{\textcolor{brown}{Problem:}} For a positive real number a, show that $10a \leq 28a^2+1$.
            \newline \textbf{\textcolor{violet}{Generated Proof:}} 
            \newline
            \texttt{have c0: $0 \leq 28*a^2 - 10*a + 1$}
            \newline
            \texttt{proof -}
            \newline
            \texttt{have c1: $(a - (5/28))^2 = a^2 - 10/28*a + (5/28)^2$ sledgehammer}
            \newline
            \texttt{then have c2: $0 \leq a^2 - 10/28*a + (5/28)^2$ using c1 sledgehammer}
            \newline
            \texttt{then have c3: $0 \leq 28*a^2 - 10*a + 28*((5/28)^2)$ using c2 sledgehammer}
            \newline
            \texttt{then have c4: $0 \leq 28*a^2 - 10*a + \textcolor{red}{5}$ using c3 sledgehammer}
            \newline
            \texttt{\textcolor{red}{then show ?thesis using c4 sledgehammer}}
            \newline
            \texttt{qed}
            \newline
            \texttt{then show ?thesis sledgehammer}
            \newline
            \texttt{qed}}\\
	    \bottomrule
		\end{tabular}
		\caption{\label{tab:errors_isabelle} Illustrations of errors made by \gptbig{} while learning Isabelle. In the top example, the model extrapolates its understanding of the keywords in Isabelle that were explained in the prompt and uses a similar word, `\texttt{hence}', which is not a part of the language. The bottom example illustrates an error in reasoning where the model makes a mistake in calculation which invalidates the statement it tries to prove.}
	}
\end{table*}
\setlength{\extrarowheight}{0pt}
\section{Example Prompts}

We provide examples of prompts for the diverse scenarios we study in Figures \ref{fig:prompt_constrained_demos}, \ref{fig:prompt_constrained_desc}, \ref{fig:prompt_isabelle_demos}, and \ref{fig:prompt_isabelle_desc_alias}.

\begin{figure*}
    \centering
    \small
\begin{lstlisting}[breaklines]

You are given the following programming problem.

You are given a string s consisting of n lowercase Latin letters. Let's define a substring as a 
contiguous subsegment of a string. For example, "acab" is a substring of "abacaba" (it starts in 
position 3 and ends in position 6), but "aa" or "d" aren't substrings of this string. 
[...truncated...] 
The lexicographic comparison of strings is implemented by operator $<$ in modern programming 
languages.

Input
The first line contains one integer n (2 $\leq$ n $\leq$ 3 $\times$ 10^5) - the length of s.
The second line contains the string s of length n consisting only of lowercase Latin letters.

Output
If it is impossible to reverse some substring of the given string to obtain a string which is 
lexicographically less, print "NO". Otherwise print "YES" and two indices l and r (1 $\leq$ l 
$<$ r $\leq$ n) denoting the substring you have to reverse. If there are multiple answers, you can 
print any.

Examples:

Input
7
abacaba

Output
YES
2 5

You are given the following novel API functions already implemented. In your code, you must 
compulsorily call these functions.

Function: xxqev()
Examples using the function xxqev():

Instruction: Add the string 'apple' into the heap of fruits, maintaining the order of the heap.
Code:
fruits = ['banana', 'cherry', 'date']
xxqev(fruits, 'apple')

Instruction: [...truncated...]

Function: adygr()
Examples using the function adygr():

Instruction: From the list of student names, pick a random student.
Code:
students = ['John', 'Amy', 'Peter', 'Anna', 'Mike']
adygr(students)

Instruction: [...truncated...]

The solution code that uses the novel API functions xxqev() and adygr():
\end{lstlisting}
    \caption{Example prompt for constrained generation from demonstrations of functions provided in-context.}
    \label{fig:prompt_constrained_demos}
\end{figure*}

\begin{figure*}
    \centering
    \small
\begin{lstlisting}[breaklines]

You are given the following programming problem.

In late autumn evening n robots gathered in the cheerful company of friends. Each robot has a unique 
identifier - an integer from 1 to 109. At some moment, robots decided to play the game "Snowball". 
Below there are the rules of this game. First, all robots stand in a row. Then the first robot says 
[...truncated...]
Your task is to determine the k-th identifier to be pronounced.

Input
The first line contains two positive integers n and k (1 $\leq$ n $\leq$ 100 000, 1 $\leq$ k $\leq$ 
min(2$\times$109, n$\times$(n + 1) / 2).
The second line contains the sequence id1, id2, ..., idn (1 $\leq$ idi $\leq$ 109) - identifiers of 
robots. It is guaranteed that all identifiers are different.

Output
Print the k-th pronounced identifier (assume that the numeration starts from 1).

Examples

Input
2 2
1 2

Output
1

You are given the following novel API functions already implemented. In your code, you must 
compulsorily call these functions.

Function: fubba()
Description: 
This function is used to calculate the square root of a given number. It accepts both positive and negative numbers as well as complex numbers. In the case of negative numbers, it returns the square root as a complex number.
Input Argument: A number (integer, float, or complex)
Return Type: complex

Function: snxlt()
Description: 
This function takes a number as an argument and returns the largest integer value less than or equal to the given number. If the input number is already an integer, the function returns the same number. It essentially rounds down the number to the nearest integer.
Input Arguments:
1. x (float or any python object that can be coerced into a float)
Return Type: Integer

Function: evpwd()
Description: 
This function takes in a floating-point number and returns the smallest integer value greater than or equal to the given number. If the number is already an integer, the same number is returned. This operation is also known as ceiling operation.
Input Arguments:
1. x (float)
Return Type: int

The solution code that uses the novel API functions fubba(), snxlt(), and evpwd():
\end{lstlisting}
    \caption{Example prompt for constrained generation from descriptions of functions provided in-context.}
    \label{fig:prompt_constrained_desc}
\end{figure*}

\begin{figure*}
    \centering
    \small
\begin{lstlisting}[breaklines]

Translate the informal solution into a sketch of the formal Isabelle proof. Add `sledgehammer` in the 
sketch whenever possible. `sledgehammer` will be used to call the automated Sledgehammer prover. 
Here are some examples:

Informal:
(*### Problem
Show that for any four complex numbers a, b, c, and d, $(a-d)(a-c)(a-b) = -(((a^2 - a(b+c)) + bc) * d) + (a^2 - a(b+c) + bc) * a$.
### Solution
We first see that $a^2 = a * a$ trivially.
Unfolding this, the main equation holds true when terms are rearranged.*)

Formal:
theorem
  fixes a b c d :: complex
  shows "(a-d) * (a-c) * (a-b) = -(((a^2 - (b+c) * a) + c * b) * d) + (a^2 - (b+c) * a + c * b) * a"
proof -
  (* We first see that $a^2 = a * a$ trivially. *)
  have t0: "a^2 = a * a"
    using power2_eq_square
      sledgehammer
  (* Unfolding this, the main equation holds true when terms are rearranged. *)
  show ?thesis unfolding t0
    sledgehammer
qed

.
.
.

Informal:
(*### Problem
Find the minimum value of $\frac{9x^2\sin^2 x + 4}{x\sin x}$ for $0 < x < \pi$. Show that it is 12.
### Solution
Let $y = x \sin x$. It suffices to show that $12 \leq \frac{9y^2 + 4}{y}.
It is trivial to see that $y > 0$.
Then one can multiply both sides by $y$ and it suffices to show $12y \leq 9y^2 + 4$.
This can be done by the sum of squares method.*)

Formal:
theorem
  fixes x::real
  assumes "0<x" "x<pi"
  shows "12 \<le> ((9 * (x^2 * (sin x)^2)) + 4) / (x * sin x)"
proof -
  define y where "y=x * sin x"
  (* It suffices to show that $12 \leq \frac{9y^2 + 4}{y}. *)
  have "12 \<le> (9 * y^2 + 4) / y"
  ---[truncated]---
  then show ?thesis
    sledgehammer
qed

Informal:
(*### Problem
What is the positive difference between $120\%$ of 30 and $130\%$ of 20? Show that it is 10.
### Solution
One hundred twenty percent of 30 is $120\cdot30\cdot\frac{1}{100}=36$, and $130\%$ of 20 is $ 130\cdot 20\cdot\frac{1}{100}=26$.  The difference between 36 and 26 is $10$.*)

Formal:
theorem
  "abs ((120::real) / 100 * 30 - 130 / 100 * 20) = 10"
proof -
\end{lstlisting}
    \caption{Example prompt for learning Isabelle from demonstrations (without aliasing).}
    \label{fig:prompt_isabelle_demos}
\end{figure*}

\begin{figure*}
    \centering
    \small
\begin{lstlisting}[breaklines]

I am defining a new programming language that you must use to prove the theorem statement provided to 
you.

The keywords of the language are:

"alyckwm": This keyword is used to declare the start of a theorem. The name of the theorem follows 
this keyword. Theorems represent mathematical truths.

"lpicd": This keyword is used to declare a variable in the context of the current theorem.

"jepxahf": This keyword is used to declare the assumptions of the theorem. Assumptions are premises 
or logical statements that are taken as given for proving the theorem.

"kdi": This keyword is used to chain multiple assumptions or multiple goals together in a single 
statement. In the context of assumptions, this allows you to state several assumptions in a 
single "jepxahf" block. Each assumption will be separated by this keyword and can be referred to 
individually in the proof. In the context of goals, if you have multiple things to prove, you 
can use this to state all these goals at once in an "fwozy" block. Then, in the proof, you can 
prove each goal one by one. The proof is only complete when all goals have been proven.

"fwozy": This keyword is used to declare the conclusion of the theorem. It represents the statement 
that needs to be proved.

"becj" : This keyword is used to establish intermediate facts within the proof. It essentially allows 
us to break down the proof into smaller, more manageable pieces. After this keyword, you state 
some logical statement that you believe to be true and that you want to prove. The proof of this 
statement is then provided after this line. You should support this statement with previous 
hypothesis specified by the "using" keyword explained below.

"curja": This keyword is used to specify the assumptions or previously proved facts that will be used 
to prove the current fact.

"hiwhrxplibce" : This is a powerful automatic proof tool or method that attempts to prove the 
statement immediately preceding it. It is often used after "becj" to prove the intermediate fact. 
It uses a variety of automated reasoning tools to find the proof.

"yeha" : This keyword is used to refer to the fact that was just proven. It allows you to use the 
result of one line in the next line of the proof.

"oiql" : This keyword is similar to "becj", but is used to prove the final goal or thesis. It is used 
to state what we want to prove and is followed by the proof of this statement.

"tnkdpfe" : This is a special term that refers to the statement of the theorem we are trying to 
prove. In other words, it's a placeholder for the goal of the proof. You do not need to state the 
logical statement again after this.

"srf" : This keyword is used to end a proof or sub-proof.

Logical statements are enclosed in double quotes such as "abs ((120::real) / 100 * 30 - 130 / 100 * 
20) = 10". Logical statements can be assigned a name such as "h0" or "h1" for instance when they 
are used with keywords such as "jepxahf", etc. For instance: becj h2 : "<statement>" curja h0 h1.

Now, using only the above keywords provided to you, prove the following theorem. Do not worry about 
syntax. 

theorem
  "abs ((120::real) / 100 * 30 - 130 / 100 * 20) = 10"

I am providing an informal proof sketch for your reference:
One hundred twenty percent of 30 is $120\cdot30\cdot\frac{1}{100}=36$, and $130\%$ of 20 is $ 130
\cdot 20\cdot\frac{1}{100}=26$.  The difference between 36 and 26 is $10$.
\end{lstlisting}
    \caption{Example prompt for learning Isabelle from description of keywords (with aliasing).}
    \label{fig:prompt_isabelle_desc_alias}
\end{figure*}

\end{document}